# Markov Localization for Mobile Robots
# in Dynamic Environments

**Dieter Fox**                                          DFOX@CS.CMU.EDU
*Computer Science Department and Robotics Institute*
*Carnegie Mellon University*
*Pittsburgh, PA 15213-3891*

**Wolfram Burgard**                      BURGARD@INFORMATIK.UNI-FREIBURG.DE
*Department of Computer Science*
*University of Freiburg*
*D-79110 Freiburg, Germany*

**Sebastian Thrun**                                    THRUN@CS.CMU.EDU
*Computer Science Department and Robotics Institute*
*Carnegie Mellon University*
*Pittsburgh, PA 15213-3891*

## Abstract

Localization, that is the estimation of a robot's location from sensor data, is a fundamental problem in mobile robotics. This papers presents a version of Markov localization which provides accurate position estimates and which is tailored towards dynamic environments. The key idea of Markov localization is to maintain a probability density over the space of all locations of a robot in its environment. Our approach represents this space metrically, using a fine-grained grid to approximate densities. It is able to globally localize the robot from scratch and to recover from localization failures. It is robust to approximate models of the environment (such as occupancy grid maps) and noisy sensors (such as ultrasound sensors). Our approach also includes a filtering technique which allows a mobile robot to reliably estimate its position even in densely populated environments in which crowds of people block the robot's sensors for extended periods of time. The method described here has been implemented and tested in several real-world applications of mobile robots, including the deployments of two mobile robots as interactive museum tour-guides.

## 1. Introduction

Robot localization has been recognized as one of the most fundamental problems in mobile robotics (Cox & Wilfong, 1990; Borenstein *et al.*, 1996). The aim of localization is to estimate the postition of a robot in its environment, given a map of the environment and sensor data. Most successful mobile robot systems to date utilize localization, as knowledge of the robot's position is essential for a broad range of mobile robot tasks.

Localization—often referred to as *position estimation* or *position control*—is currently a highly active field of research, as a recent book by Borenstein and colleagues (1996) suggests. The localization techniques developed so far can be distinguished according to the type of problem they attack. *Tracking* or *local* techniques aim at compensating odometric errors occurring during robot navigation. They require, however, that the initial location of the robot is (approximately) known and they typically cannot recover if they lose track of the robot's position (within certain bounds). Another family of approaches is called *global* techniques. These are designed to estimate the position of the robot even under global uncertainty. Techniques of this type solve the so-called *wake-up robot problem*, in that they can localize a robot without any prior knowledge about its position. They furthermore can handle the *kidnapped robot problem*, in which a robot is carried to an arbitrary location during it's operation[1]. Global localization techniques are more powerful than local ones. They typically can cope with situations in which the robot is likely to experience serious positioning errors.

In this paper we present a metric variant of Markov localization, a technique to globally estimate the position of a robot in its environment. Markov localization uses a probabilistic framework to maintain a position probability density over the whole set of possible robot poses. Such a density can have arbitrary forms representing various kinds of information about the robot's position. For example, the robot can start with a uniform distribution representing that it is completely uncertain about its position. It furthermore can contain multiple modes in the case of ambiguous situations. In the usual case, in which the robot is highly certain about its position, it consists of a unimodal distribution centered around the true position of the robot. Based on the probabilistic nature of the approach and the representation, Markov localization can globally estimate the position of the robot, it can deal with ambiguous situations, and it can re-localize the robot in the case of localization failures. These properties are basic preconditions for truly autonomous robots designed to operate over long periods of time.

Our method uses a fine-grained and metric discretization of the state space. This approach has several advantages over previous ones, which predominately used Gaussians or coarse-grained, topological representations for approximating a robot's belief. First, it provides more accurate position estimates, which are required in many mobile robot tasks (e.g., tasks involving mobile manipulation). Second, it can incorporate raw sensory input such as a single beam of an ultrasound sensor. Most previous approaches to Markov localization, in contrast, screen sensor data for the presence or absence of landmarks, and they are prone to fail if the environment does not align well with the underlying assumptions (e.g., if it does not contain any of the required landmarks).

Most importantly, however, previous Markov localization techniques assumed that the environment is *static*. Therefore, they typically fail in highly dynamic environments, such as public places where crowds of people may cover the robot's sensors for extended periods of time. To deal with such situations, our method applies a filtering technique that, in essence, updates the position probability density using only those measurements which are with high likelihood produced by known objects contained in the map. As a result, it permits accurate localization even in densely crowded, non-static environments.

Our Markov localization approach has been implemented and evaluated in various environments, using different kinds of robots and sensor modalities. Among these applications are the deployments of the mobile robots Rhino and Minerva (see Figure 1) as interac-tive museum tour-guide robots (Burgard *et al.*, 1998a, 2000; Thrun *et al.*, 1999) in the *Deutsches Museum Bonn* and the *National Museum of American History* in Washington, DC, respectively. Experiments described in this paper illustrate the ability of our Markov localization technique to deal with approximate models of the environment, such as occupancy grid maps and noisy sensors such as ultrasound sensors, and they demonstrate that our approach is well-suited to localize robots in densely crowded environments, such as museums full of people.

---

1. Please note that the wake-up problem is the special case of the kidnapped robot problem in which the robot is told that it has been carried away.

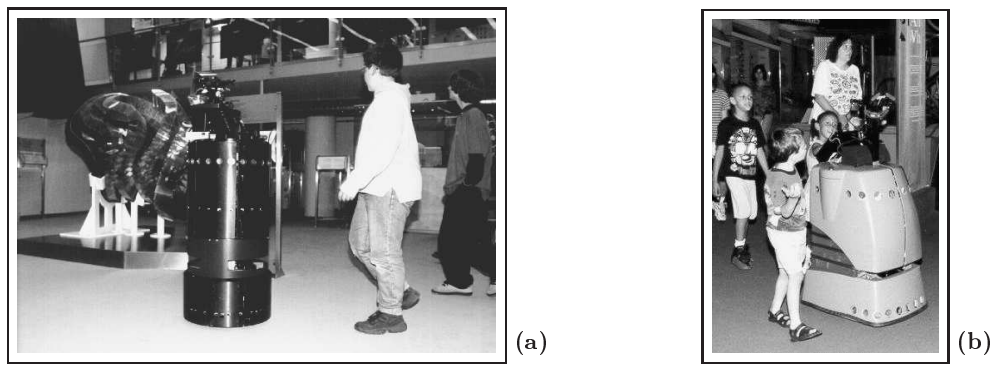

Fig. 1. The mobile robots Rhino (a) and Minerva (b) acting as interactive museum tour-guides.

The paper is organized as follows. The next section describes the mathematical framework of Markov localization. We introduce our metric version of Markov localization in Section 3. This section also presents a probabilistic model of proximity sensors and a filtering scheme to deal with highly dynamic environments. Thereafter, we describe experimental results illustrating different aspects of our approach. Related work is discussed in Section 5 followed by concluding remarks.

## 2. Markov Localization

To introduce the major concepts, we will begin with an intuitive description of Markov localization, followed by a mathematical derivation of the algorithm. The reader may notice that Markov localization is a special case of probabilistic state estimation, applied to mobile robot localization (see also Russell & Norvig, 1995; Fox, 1998 and Koenig & Simmons, 1998).

For clarity of the presentation, we will initially make the restrictive assumption that the environment is *static*. This assumption, called *Markov assumption*, is commonly made in the robotics literature. It postulates that the robot's location is the only state in the environment which systematically affects sensor readings. The Markov assumption is violated if robots share the same environment with people. Further below, in Section 3.3, we will side-step this assumption and present a Markov localization algorithm that works well even in highly dynamic environments, e.g., museums full of people.

### 2.1 The Basic Idea

Markov localization addresses the problem of state estimation from sensor data. Markov localization is a probabilistic algorithm: Instead of maintaining a single hypothesis as to where in the world a robot might be, Markov localization maintains a *probability distribution* over the space of all such hypotheses. The probabilistic representation allows it to weigh these different hypotheses in a mathematically sound way.

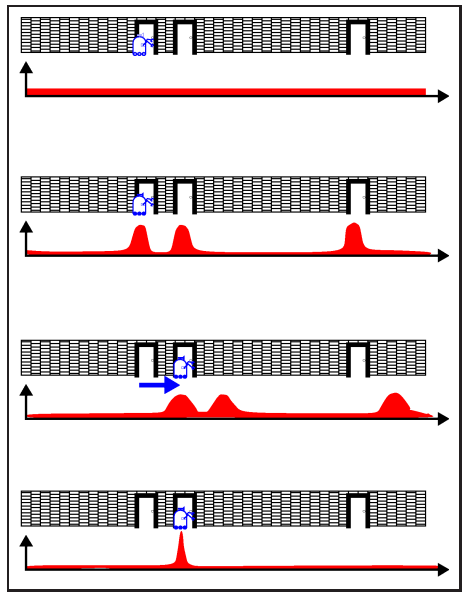

Fig. 2. The basic idea of Markov localization: A mobile robot during global localization.

Before we delve into mathematical detail, let us illustrate the basic concepts with a simple example. Consider the environment depicted in Figure 2. For the sake of simplicity, let us assume that the space of robot positions is one-dimensional, that is, the robot can only move horizontally (it may not rotate). Now suppose the robot is placed somewhere in this environment, but it is not told its location. Markov localization represents this state of uncertainty by a *uniform distribution* over all positions, as shown by the graph in the first diagram in Figure 2. Now let us assume the robot queries its sensors and finds out that it is next to a door. Markov localization modifies the belief by raising the probability for places next to doors, and lowering it anywhere else. This is illustrated in the second diagram in Figure 2. Notice that the resulting belief is multi-modal, reflecting the fact that the available information is insufficient for global localization. Notice also that places not next to a door still possess non-zero probability. This is because sensor readings are noisy, and a single sight of a door is typically insufficient to exclude the possibility of not being next to a door.

Now let us assume the robot moves a meter forward. Markov localization incorporates this information by shifting the belief distribution accordingly, as visualized in the third diagram in Figure 2. To account for the inherent noise in robot motion, which inevitably leads to a loss of information, the new belief is smoother (and less certain) than the previous one. Finally, let us assume the robot senses a second time, and again it finds itself next to a door. Now this observation is multiplied into the current (non-uniform) belief, which leads to the final belief shown at the last diagram in Figure 2. At this point in time, most of the probability is centered around a single location. The robot is now quite certain about its position.

## 2.2 Basic Notation

To make this more formal, let us denote the *position* (or: *location*) of a mobile robot by a three-dimensional variable $l = \langle x, y, \theta \rangle$, comprising its $x$-$y$ coordinates (in some Cartesian coordinate system) and its heading direction $\theta$. Let $l_t$ denote the robot's true location at time $t$, and $L_t$ denote the corresponding random variable. Throughout this paper, we will use the terms position and location interchangeably.

Typically, the robot does not know its exact position. Instead, it carries a belief as to where it might be. Let $Bel(L_t)$ denote the robot's position belief at time $t$. $Bel(L_t)$ is a probability distribution over the space of positions. For example, $Bel(L_t = l)$ is the probability (density) that the robot assigns to the possibility that its location at time $t$ is $l$. The belief is updated in response to two different types of events: The arrival of a measurement through the robot's environment sensors (e.g., a camera image, a sonar scan), and the arrival of an odometry reading (e.g., wheel revolution count). Let us denote environment sensor measurements by $s$ and odometry measurements by $a$, and the corresponding random variables by $S$ and $A$, respectively.

The robot perceives a stream of measurements, sensor measurements $s$ and odometry readings $a$. Let

$$d = \{d_0, d_1, \ldots, d_T\} \tag{1}$$

denote the stream of measurements, where each $d_t$ (with $0 \leq t \leq T$) either is a sensor measurement or an odometry reading. The variable $t$ indexes the data, and $T$ is the most recently collected data item (one might think of $t$ as "time"). The set $d$, which comprises all available sensor data, will be referred to as the *data*.

## 2.3 Recursive Localization

Markov localization estimates the posterior distribution over $L_T$ conditioned on all available data, that is

$$P(L_T = l \mid d) = P(L_T = l \mid d_0, \ldots, d_T). \tag{2}$$

Before deriving incremental update equations for this posterior, let us briefly make explicit the key assumption underlying our derivation, called the *Markov assumption*. The Markov assumption, sometimes referred to as *static world assumption*, specifies that if one knows the robot's location $l_t$, future measurements are independent of past ones (and vice versa):

$$P(d_{t+1}, d_{t+2}, \ldots \mid L_t = l, d_0, \ldots, d_t) = P(d_{t+1}, d_{t+2}, \ldots \mid L_t = l) \ \forall t \tag{3}$$

In other words, we assume that the robot's location is the only state in the environment, and knowing it is all one needs to know about the past to predict future data. This assumption is clearly inaccurate if the environment contains moving (and measurable) objects other than the robot itself. Further below, in Section 3.3, we will extend the basic paradigm to non-Markovian environments, effectively devising a localization algorithm that works well in a broad range of dynamic environments. For now, however, we will adhere to the Markov assumption, to facilitate the derivation of the basic algorithm.

When computing $P(L_T = l \mid d)$, we distinguish two cases, depending on whether the most recent data item $d_T$ is a sensor measurement or an odometry reading.

**Case 1: The most recent data item is a sensor measurement $d_T = s_T$.**

Here

$$P(L_T = l \mid d) \quad = \quad P(L_T = l \mid d_0, \ldots, d_{T-1}, s_T). \tag{4}$$

Bayes rule suggests that this term can be transformed to

$$\frac{P(s_T \mid d_0, \ldots, d_{T-1}, L_T = l) \; P(L_T = l \mid d_0, \ldots, d_{T-1})}{P(s_T \mid d_0, \ldots, d_{T-1})}, \tag{5}$$

which, because of our Markov assumption, can be simplified to:

$$\frac{P(s_T \mid L_T = l) \; P(L_T = l \mid d_0, \ldots, d_{T-1})}{P(s_T \mid d_0, \ldots, d_{T-1})}. \tag{6}$$

We also observe that the denominator can be replaced by a constant $\alpha_T$, since it does not depend on $L_T$. Thus, we have

$$P(L_T = l \mid d) \quad = \quad \alpha_T \; P(s_T \mid L_T = l) \; P(L_T = l \mid d_0, \ldots, d_{T-1}). \tag{7}$$

The reader may notice the incremental nature of Equation (7): If we write

$$Bel(L_T = l) \quad = \quad P(L_T = l \mid d_0, \ldots, d_T), \tag{8}$$

to denote the robot's belief Equation (7) becomes

$$Bel(L_T = l) \quad = \quad \alpha_T \; P(s_T \mid l) \; Bel(L_{T-1} = l). \tag{9}$$

In this equation we replaced the term $P(s_T \mid L_T = l)$ by $P(s_T \mid l)$ based on the assumption that it is independent of the time.

**Case 2: The most recent data item is an odometry reading: $d_T = a_T$.**

Here we compute $P(L_T = l \mid d)$ using the Theorem of Total Probability:

$$P(L_T = l \mid d) \quad = \quad \int P(L_T = l \mid d, L_{T-1} = l') \; P(L_{T-1} = l' \mid d) \; dl'. \tag{10}$$

Consider the first term on the right-hand side. Our Markov assumption suggests that

$$P(L_T = l \mid d, L_{T-1} = l') \quad = \quad P(L_T = l \mid d_0, \ldots, d_{T-1}, a_T, L_{T-1} = l') \tag{11}$$

$$= \quad P(L_T = l \mid a_T, L_{T-1} = l') \tag{12}$$

The second term on the right-hand side of Equation (10) can also be simplified by observing that $a_T$ does not carry any information about the position $L_{T-1}$:

$$P(L_{T-1} = l' \mid d) \quad = \quad P(L_{T-1} = l' \mid d_0, \ldots, d_{T-1}, a_T) \tag{13}$$

$$= \quad P(L_{T-1} = l' \mid d_0, \ldots, d_{T-1}) \tag{14}$$

Substituting 12 and 14 back into Equation (10) gives us the desired result

$$P(L_T = l \mid d) \quad = \quad \int P(L_T = l \mid a_T, L_{T-1} = l') \; P(L_{T-1} = l' \mid d_0, \dots, d_{T-1}) \; dl'. \quad (15)$$

Notice that Equation (15) is, too, of an incremental form. With our definition of belief above, we have

$$Bel(L_T = l) \quad = \quad \int P(l \mid a_T, l') \; Bel(L_{T-1} = l') \; dl'. \quad (16)$$

Please note that we used $P(l \mid a_T, l')$ instead of $P(L_T = l \mid a_T, L_{T-1} = l')$ since we assume that it does not change over time.

## 2.4 The Markov Localization Algorithm

Update Equations (9) and (16) form the core of the Markov localization algorithm. The full algorithm is shown in Table 1. Following Basye et al. (1992) and Russell & Norvig (1995), we denote $P(l \mid a, l')$ as the robot's *motion model*, since it models how motion effect the robot's position. The conditional probability $P(s \mid l)$ is called *perceptual model*, because it models the outcome of the robot's sensors.

In the Markov localization algorithm $P(L_0 = l)$, which initializes the belief $Bel(L_0)$, reflects the *prior* knowledge about the starting position of the robot. This distribution can be initialized arbitrarily, but in practice two cases prevail: If the position of the robot relative to its map is entirely unknown, $P(L_0)$ is usually uniformly distributed. If the initial position of the robot is approximately known, then $P(L_0)$ is typically a narrow Gaussian distribution centered at the robot's position.

## 2.5 Implementations of Markov Localization

The reader may notice that the principle of Markov localization leaves open

1. how the robot's belief $Bel(L)$ is represented and

2. how the conditional probabilities $P(l \mid a, l')$ and $P(s \mid l)$ are computed.

Accordingly, existing approaches to Markov localization mainly differ in the representation of the state space and the computation of the perceptual model. In this section we will briefly discuss different implementations of Markov localization focusing on these two topics (see Section 5 for a more detailed discussion of related work).

1. **State Space Representations:** A very common approach for the representation of the robots belief $Bel(L)$ is based on Kalman filtering (Kalman, 1960; Smith *et al.*, 1990) which rests on the restrictive assumption that the position of the robot can be modeled by a unimodal Gaussian distribution. Existing implementations (Leonard & Durrant-Whyte, 1992; Schiele & Crowley, 1994; Gutmann & Schlegel, 1996; Arras & Vestli, 1998) have proven to be robust and accurate for keeping track of the robot's position. Because of the restrictive assumption of a Gaussian distribution these techniques lack the ability to represent situations in which the position of the robot

---

**for each** location $l$ **do**                             /* *initialize* the belief */

$$Bel(L_0 = l) \quad \longleftarrow \quad P(L_0 = l) \tag{17}$$

**end for**

**forever do**

    **if** new sensory input $s_T$ is received **do**

        $\alpha_T \longleftarrow 0$

        **for each** location $l$ **do**                    /* apply the *perception model* */

$$\widehat{Bel}(L_T = l) \quad \longleftarrow \quad P(s_T \mid l) \cdot Bel(L_{T-1} = l) \tag{18}$$

$$\alpha_T \quad \longleftarrow \quad \alpha_T + \widehat{Bel}(L_T = l) \tag{19}$$

        **end for**

        **for each** location $l$ **do**                     /* *normalize* the belief */

$$Bel(L_T = l) \quad \longleftarrow \quad \alpha_T^{-1} \cdot \widehat{Bel}(L_T = l) \tag{20}$$

        **end for**

    **end if**

    **if** an odometry reading $a_T$ is received **do**

        **for each** location $l$ **do**                    /* apply the *motion model* */

$$Bel(L_T = l) \quad \longleftarrow \quad \int P(l \mid l', a_T) \cdot Bel(L_{T-1} = l') \, dl' \tag{21}$$

        **end for**

    **end if**

**end forever**

---

Tab. 1. The Markov localization algorithm

maintains multiple, distinct beliefs (c.f. 2). As a result, localization approaches using Kalman filters typically require that the starting position of the robot is known and are not able to re-localize the robot in the case of localization failures. Additionally, Kalman filters rely on sensor models that generate estimates with Gaussian uncertainty. This assumption, unfortunately, is not met in all situations (see for example Dellaert *et al.* 1999).

To overcome these limitations, different approaches have used increasingly richer schemes to represent uncertainty in the robot's position, moving beyond the Gaussian density assumption inherent in the vanilla Kalman filter. Nourbakhsh *et al.* (1995), Simmons & Koenig (1995), and Kaelbling et al. (1996) use Markov localization for landmark-based corridor navigation and the state space is organized according to the coarse, topological structure of the environment and with generally only four possible orientations of the robot. These approaches can, in principle, solve the problem of global localization. However, due to the coarse resolution of the state representation, the accuracy of the position estimates is limited. Topological approaches typically give only a rough sense as to where the robot is. Furthermore, these techniques require that the environment satisfies an orthogonality assumption and that there are certain landmarks or abstract features that can be extracted from the sensor data. These assumptions make it difficult to apply the topological approaches in unstructured environments.

2. **Sensor Models:** In addition to the different representations of the state space various perception models have been developed for different types of sensors (see for example Moravec, 1988; Kortenkamp & Weymouth, 1994; Simmons & Koenig, 1995; Burgard *et al.*, 1996; Dellaert *et al.*, 1999; and Konolige, 1999). These sensor models differ in the way how they compute the probability of the current measurement. Whereas topological approaches such as (Kortenkamp & Weymouth, 1994; Simmons & Koenig, 1995; Kaelbling *et al.*, 1996) first extract landmark information out of a sensor scan, the approaches in (Moravec, 1988; Burgard *et al.*, 1996; Dellaert *et al.*, 1999; Konolige, 1999) operate on the raw sensor measurements. The techniques for proximity sensors described in (Moravec, 1988; Burgard *et al.*, 1996; Konolige, 1999) mainly differ in their efficiency and how they model the characteristics of the sensors and the map of the environment.

In order to combine the strengths of the previous representations, our approach relies on a fine and less restrictive representation of the state space (Burgard *et al.*, 1996, 1998b; Fox, 1998). Here the robot's belief is approximated by a fine-grained, regularly spaced grid, where the spatial resolution is usually between 10 and 40 cm and the angular resolution is usually 2 or 5 degrees. The advantage of this approach compared to the Kalman-filter based techniques is its ability to represent multi-modal distributions, a prerequisite for global localization from scratch. In contrast to the topological approaches to Markov localization, our approach allows accurate position estimates in a much broader range of environments, including environments that might not even possess identifiable landmarks. Since it does not depend on abstract features, it can incorporate raw sensor data into the robot's belief. And it typically yields results that are an order of magnitude more accurate. An obvious shortcoming of the grid-based representation, however, is the size of the state space that has to be maintained. Section 3.4 addresses this issue directly by introducing techniques that make it possible to update extremely large grids in real-time.

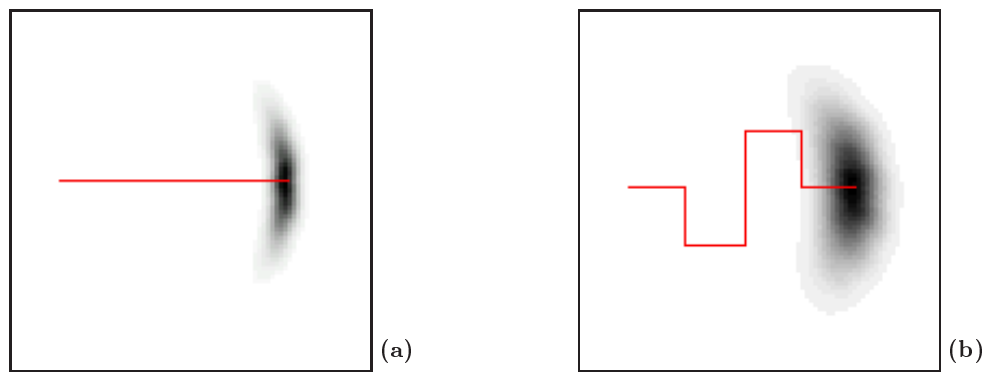

Fig. 3. Typical "banana-shaped" distributions resulting from different motion actions.

## 3. Metric Markov Localization for Dynamic Environments

In this section we will describe our metric variant of Markov localization. This includes appropriate motion and sensor models. We also describe a filtering technique which is designed to overcome the assumption of a static world model generally made in Markov localization and allows to localize a mobile robot even in densely crowded environments. We then describe our fine-grained grid-based representation of the state space and present techniques to efficiently update even large state spaces.

### 3.1 The Action Model

To update the belief when the robot moves, we have to specify the action model $P(l \mid l', a_t)$. Based on the assumption of normally distributed errors in translation and rotation, we use a mixture of two independent, zero-centered Gaussian distributions whose tails are cut off (Burgard $et$ $al.$, 1996). The variances of these distributions are proportional to the length of the measured motion.

Figure 3 illustrates the resulting densities for two example paths if the robot's belief starts with a Dirac distribution. Both distributions are three-dimensional (in $\langle x, y, \theta \rangle$-space) and Figure 3 shows their 2D projections into $\langle x, y \rangle$-space.

### 3.2 The Perception Model for Proximity Sensors

As mentioned above, the likelihood $P(s \mid l)$ that a sensor reading $s$ is measured at position $l$ has to be computed for all positions $l$ in each update of the Markov localization algorithm (see Table 1). Therefore, it is crucial for on-line position estimation that this quantity can be computed very efficiently. Moravec (1988) proposed a method to compute a generally non-Gaussian probability density function $P(s \mid l)$ over a discrete set of possible distances measured by an ultrasound sensor at location $l$. In a first implementation of our approach (Burgard $et$ $al.$, 1996) we used a similar method, which unfortunately turned out to be computationally too expensive for localization in real-time.

To overcome this disadvantage, we developed a sensor-model which allows to compute $P(s \mid l)$ solely based on the distance $o_l$ to the closest obstacle in the map along the direction of the sensor. This distance can be computed by ray-tracing in occupancy grid maps or

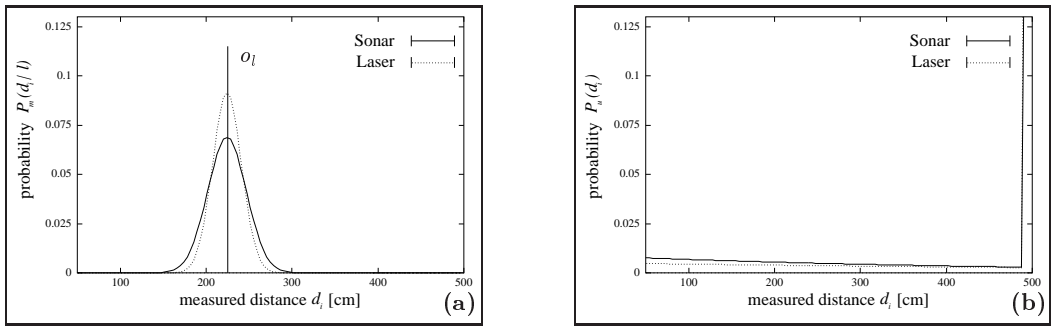

Fig. 4. Probability of measuring a distance $d_i$ (a) if obstacle in distance $o_l$ is detected and (b) due to unknown obstacles.

CAD-models of the environment. In particular, we consider a discretization $d_1, \ldots, d_n$ of possible distances measured by a proximity sensor. In our discretization, the size of the ranges $\Delta_d = d_{i+1} - d_i$ is the same for all $i$, and $d_n$ corresponds to the maximal range of the proximity sensor[2]. Let $P(d_i \mid l)$ denote the probability of measuring a distance $d_i$ if the robot is at location $l$. In order to derive this probability we first consider the following two cases (see also Hennig 1997 and Fox 1998):

a.) **Known obstacles**: If the sensor detects an obstacle the resulting distribution is modeled by a Gaussian distribution with mean at the distance to this obstacle. Let $P_m(d \mid l)$ denote the probability of measuring distance $d$ if the robot is at location $l$, assuming that the sensor beam is reflected by the closest obstacle in the map (along the sensor beam). We denote the distance to this specific obstacle by $o_l$. The probability $P_m(d \mid l)$ is then given by a Gaussian distribution with mean at $o_l$:

$$P_m(d \mid l) \;=\; \frac{1}{\sigma\sqrt{2\pi}} e^{-\frac{(d-o_l)^2}{2\sigma^2}} \qquad (22)$$

The standard deviation $\sigma$ of this distribution models the uncertainty of the measured distance, based on

- the granularity of the discretization of $L$, which represents the robot's position,
- the accuracy of the world model, and
- the accuracy of the sensor.

Figure 4(a) gives examples of such Gaussian distributions for ultrasound sensors and laser range-finders. Here the distance $o_l$ to the closest obstacle is 230cm. Observe here that the laser sensor has a higher accuracy than the ultrasound sensor, as indicated by the smaller variance.

b.) **Unknown obstacles**: In Markov localization, the world model generally is assumed to be static and complete. However, mobile robot environments are often populated and therefore contain objects that are not included in the map. Consequently, there is a non-zero probability that the sensor is reflected by an obstacle not represented in the world model. Assuming that these objects are *equally distributed* in the environment, the probability $P_u(d_i)$ of detecting an unknown obstacle at distance $d_i$ is independent of the location of the robot and can be modeled by a geometric distribution. This distribution results from the following observation. A distance $d_i$ is measured if the sensor is *not* reflected by an obstacle at a shorter distance $d_{j<i}$ *and* is reflected at distance $d_i$. The resulting probability is

$$P_u(d_i) \;=\; \left\{ \begin{array}{ll} 0 & i = 0 \\ c_r(1 - \sum_{j<i} P_u(d_j)) & \text{otherwise.} \end{array} \right. \tag{23}$$

---

2. Typical values for $n$ are between 64 and 256 and the maximal range $d_n$ is typically 500cm or 1000cm.

In this equation the constant $c_r$ is the probability that the sensor is reflected by an unknown obstacle at any range given by the discretization.

A typical distribution for sonar and laser measurements is depicted in Figure 4(b). In this example, the relatively large probability of measuring 500cm is due to the fact that the maximum range of the proximity sensors is set to 500cm. Thus, this distance represents the probability of measuring *at least* 500cm.

Obviously, only *one* of these two cases can occur at a certain point in time, i.e., the sensor beam is either reflected by a known or an unknown object. Thus, $P(d_i \mid l)$ is a a mixture of the two distributions $P_m$ and $P_u$. To determine the combined probability $P(d_i \mid l)$ of measuring a distance $d_i$ if the robot is at location $l$ we consider the following two situations: A distance $d_i$ is measured, if

a.) the sensor beam is

    1.) **not** reflected by an unknown obstacle before reaching distance $d_i$

$$a_1 = 1 - \sum_{j<i} P_u(d_j), \tag{24}$$

    2.) **and** reflected by the known obstacle at distance $d_i$

$$a_2 = c_d \, P_m(d_i \mid l) \tag{25}$$

b.) **OR** the beam is

    1.) reflected **neither** by an unknown obstacle **nor** by the known obstacle before reaching distance $d_i$

$$b_1 = 1 - \sum_{j<i} P(d_j \mid l) \tag{26}$$

    2.) **and** reflected by an unknown obstacle at distance $d_i$

$$b_2 = c_r. \tag{27}$$

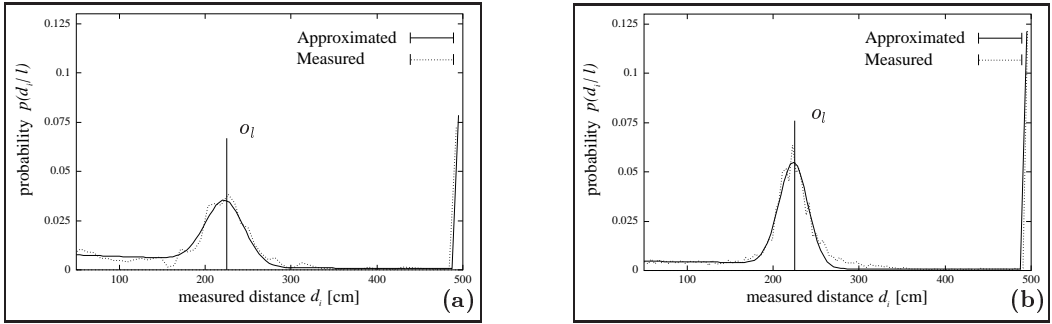

Fig. 5. Measured and approximated probabilities of (a) sonar and (b) laser measurements given the distance $o_l$ to the closest obstacle along the sensing direction.

The parameter $c_d$ in Equation (25) denotes the probability that the sensor detects the closest obstacle in the map. These considerations for the combined probability are summarized in Equation (28). By double negation and insertion of the Equations (24) to (27), we finally get Equation (31).

$$P(d_i \mid l) = p\Big( \quad (a_1 \wedge a_2) \quad \vee \quad (b_1 \wedge b_2) \quad \Big) \tag{28}$$

$$= \neg p\Big( \neg(a_1 \wedge a_2) \quad \wedge \quad \neg(b_1 \wedge b_2) \quad \Big) \tag{29}$$

$$= 1 - \Big( [1 - P(a_1 a_2)] \cdot [1 - P(b_1 b_2)] \Big) \tag{30}$$

$$= 1 - \Big( 1 - (1 - \sum_{j<i} P_u(d_j)) \, c_d \, P_m(d_i \mid l)) \Big) \cdot (1 - (1 - \sum_{j<i} P(d_j)) \, c_r) \tag{31}$$

To obtain the probability of measuring $d_n$, the maximal range of the sensor, we exploit the following equivalence: The probability of measuring a distance larger than or equal to the maximal sensor range is equivalent to the probability of *not* measuring a distance shorter than $d_n$. In our incremental scheme, this probability can easily be determined:

$$P(d_n \mid l) = 1 - \sum_{j<n} P(d_j \mid l) \tag{32}$$

To summarize, the probability of sensor measurements is computed incrementally for the different distances starting at distance $d_1 = 0$cm. For each distance we consider the probability that the sensor beam reaches the corresponding distance and is reflected either by the closest obstacle in the map (along the sensor beam), or by an unknown obstacle.

In order to adjust the parameters $\sigma$, $c_r$ and $c_d$ of our perception model we collected eleven million data pairs consisting of the expected distance $o_l$ and the measured distance $d_i$ during the typical operation of the robot. From these data we were able to estimate the probability of measuring a certain distance $d_i$ if the distance $o_l$ to the closest obstacle in the map along the sensing direction is given. The dotted line in Figure 5(a) depicts this probability for sonar measurements if the distance $o_l$ to the next obstacle is 230cm. Again, the high probability of measuring 500cm is due to the fact that this distance represents the probability of measuring *at least* 500cm. The solid line in the figure represents the distribution obtained by adapting the parameters of our sensor model so as to best fit the measured data. The corresponding measured and approximated probabilities for the laser sensor are plotted in Figure 5(b).

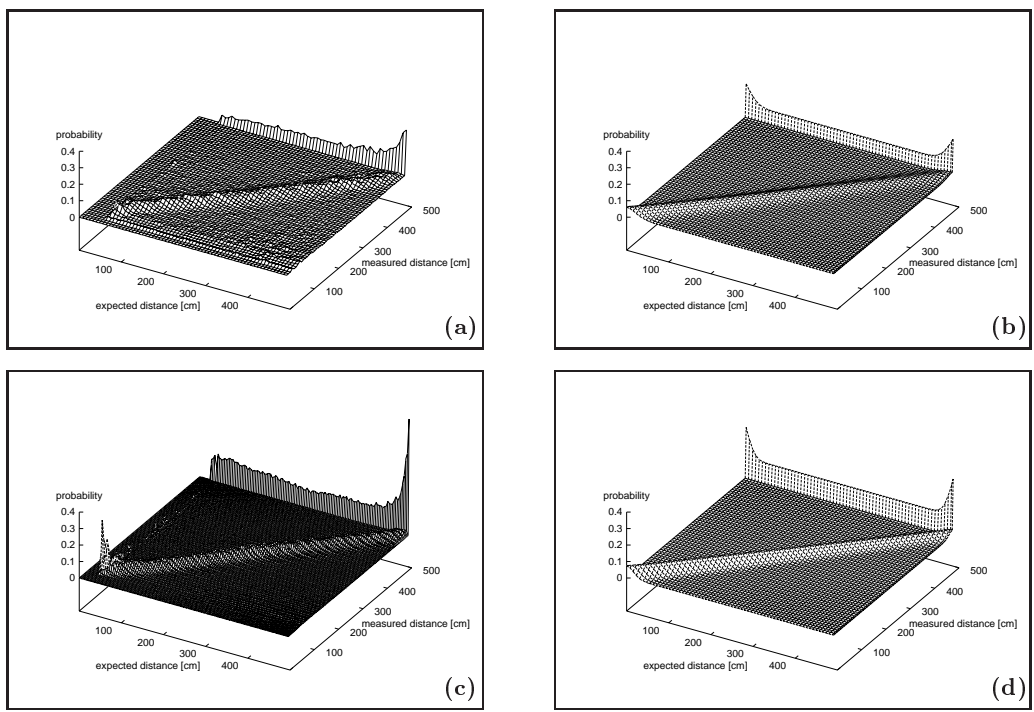

Fig. 6. Measured and approximated probability of sonar (a,b) and laser (c,d) measurements, respectively. Each table contains the probabilities of distance measurements given the expected distance $o_l$ extracted from a map of the environment.

The observed densities for *all* possible distances $o_l$ to an obstacle for ultrasound sensors and laser range-finder are depicted in Figure 6(a) and Figure 6(c), respectively. The approximated densities are shown in Figure 6(b) and Figure 6(d). In all figures, the distance $o_l$ is labeled "expected distance". The similarity between the measured and the approximated distributions shows that our sensor model yields a good approximation of the data.

Please note that there are further well-known types of sensor noise which are not explicitly represented in our sensor model. Among them are specular reflections or cross-talk which are often regarded as serious sources of noise in the context of ultra-sound sensors. However, these sources of sensor noise are modeled implicitly by the geometric distribution resulting from unknown obstacles.

## 3.3 Filtering Techniques for Dynamic Environments

Markov localization has been shown to be robust to occasional changes of an environment such as opened / closed doors or people walking by. Unfortunately, it fails to localize a robot if too many aspects of the environment are not covered by the world model. This is the case, for example, in densely crowded environments, where groups of people cover the robots sensors and thus lead to many unexpected measurements. The mobile robots Rhino and Minerva, which were deployed as interactive museum tour-guides (Burgard *et al.*, 1998a, 2000; Thrun *et al.*, 1999), were permanently faced with such a situation. Figure 7 shows two cases in which the robot Rhino is surrounded by many visitors while giving a tour in the *Deutsches Museum Bonn*, Germany.

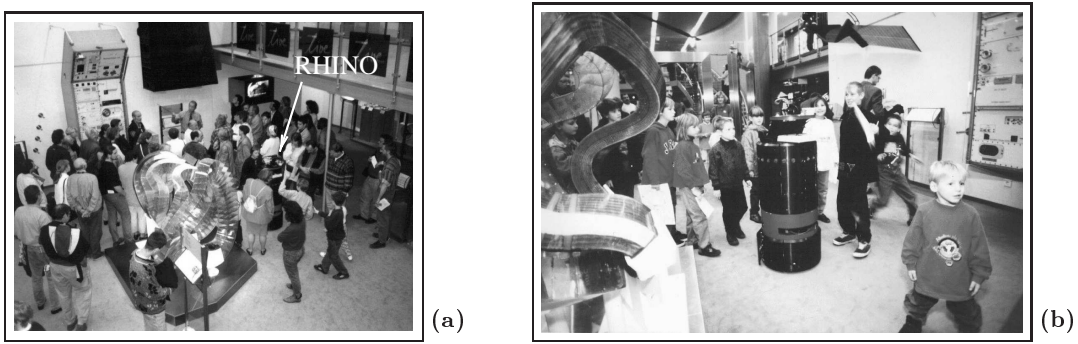

Fig. 7. Rhino surrounded by visitors in the *Deutsches Museum Bonn*.

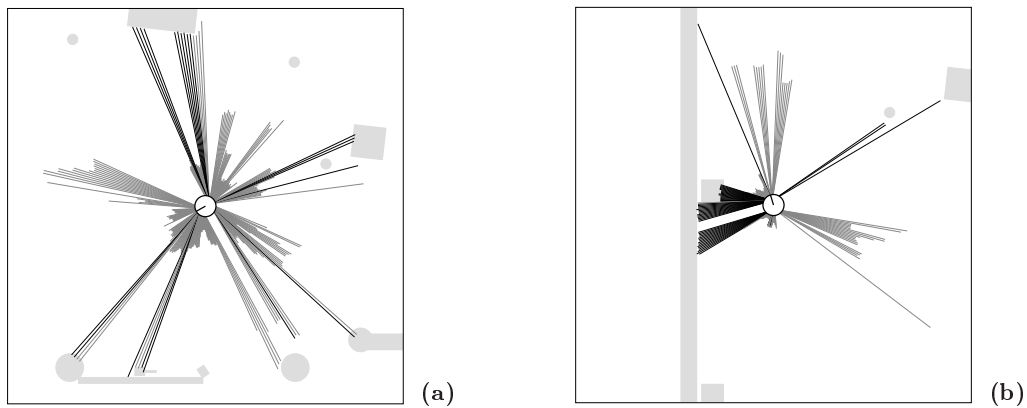

Fig. 8. Typical laser scans obtained when Rhino is surrounded by visitors.

The reason why Markov localization fails in such situations is the violation of the *Markov assumption*, an independence assumption on which virtually all localization techniques are based. As discussed in Section 2.3, this assumption states that the sensor measurements observed at time $t$ are independent of all other measurements, given that the current state $L_t$ of the world is known. In the case of localization in densely populated environments, this independence assumption is clearly violated when using a static model of the world.

To illustrate this point, Figure 8 depicts two typical laser scans obtained during the museum projects (maximal range measurements are omitted). The figure also shows the obstacles contained in the map. Obviously, the readings are, to a large extent, corrupted, since people in the museum are not represented in the static world model. The different shading of the beams indicates the two classes they belong to: the black lines correspond to the static obstacles in the map and are independent of each other if the position of the robot is known. The grey-shaded lines are beams reflected by visitors in the Museum. These sensor beams cannot be predicted by the world model and therefore are not independent of each other. Since the vicinity of people usually increases the robot's belief of being close to modeled obstacles, the robot quickly loses track of its position when incorporating all

405

sensor measurements. To reestablish the independence of sensor measurements we could include the position of the robot *and* the position of people into the state variable $L$. Unfortunately, this is infeasible since the computational complexity of state estimation increases exponentially in the number of dependent state variables to be estimated.

A closely related solution to this problem could be to adapt the map according to the changes of the environment. Techniques for concurrent map-building and localization such as (Lu & Milios, 1997a; Gutmann & Schlegel, 1996; Shatkey & Kaelbling, 1997; Thrun *et al.*, 1998b), however, also assume that the environment is almost static and therefore are unable to deal with such environments. Another approach would be to adapt the perception model to correctly reflect such situations. Note that our perceptual model already assigns a certain probability to events where the sensor beam is reflected by an unknown obstacle. Unfortunately, such approaches are only capable to model such noise *on average*. While such approaches turn out to work reliably with occasional sensor blockage, they are not sufficient in situations where more than fifty percent of the sensor measurements are corrupted. Our localization system therefore includes filters which are designed to detect whether a certain sensor reading is corrupted or not. Compared to a modification of the static sensor model described above, these filters have the advantage that they do not average over all possible situations and that their decision is based on the current belief of the robot.

The filters are designed to select those readings of a complete scan which do not come from objects contained in the map. In this section we introduce two different kinds of filters. The first one is called *entropy filter*. Since it filters a reading based solely on its effect on the belief $Bel(L)$, it can be applied to arbitrary sensors. The second filter is the *distance filter* which selects the readings according to how much shorter they are than the expected value. It therefore is especially designed for proximity sensors.

### 3.3.1 THE ENTROPY FILTER

The entropy $H(L)$ of the belief over $L$ is defined as

$$H(L) \;=\; -\sum_l Bel(L=l) \,\log Bel(L=l) \tag{33}$$

and is a measure of uncertainty about the outcome of the random variable $L$ (Cover & Thomas, 1991). The higher the entropy, the higher the robot's uncertainty as to where it is. The *entropy filter* measures the relative change of entropy upon incorporating a sensor reading into the belief $Bel(L)$. More specifically, let $s$ denote the measurement of a sensor (in our case a single range measurement). The change of the entropy of $Bel(L)$ given $s$ is defined as:

$$\Delta H(L \mid s) \;:=\; H(L \mid s) - H(L) \tag{34}$$

The term $H(L \mid s)$ is the entropy of the belief $Bel(L)$ *after* incorporating the sensor measurement $s$ (see Equations (18) – (20)). While a positive change of entropy indicates that after incorporating $s$, the robot is less certain about its position, a negative change indicates an increase in certainty. The selection scheme of the entropy filter is to exclude all sensor measurements $s$ with $\Delta H(L \mid s) < 0$. In other words, it only uses those sensor readings confirming the robot's current belief.

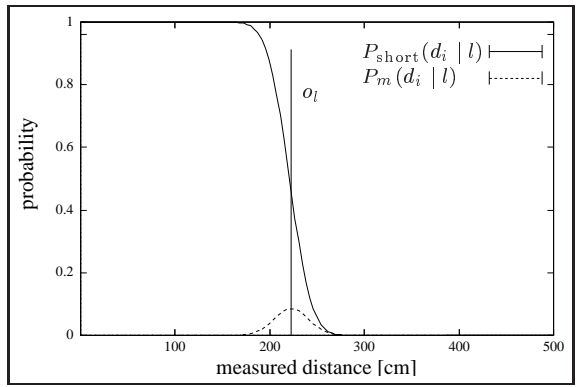

Fig. 9. Probability $P_m(d_i \mid l)$ of expected measurement and probability $P_{\text{short}}(d_i \mid l)$ that a distance $d_i$ is shorter than the expected measurement given the location $l$.

Entropy filters work well when the robot's belief is focused on the correct hypothesis. However, they may fail in situations in which the robot's belief state is incorrect. This topic will be analyzed systematically in the experiments described in Section 4.1. The advantage of the entropy filter is that it makes no assumptions about the nature of the sensor data and the kind of disturbances occurring in dynamic environments.

### 3.3.2 The Distance Filter

The distance filter has specifically been designed for proximity sensors such as laser range-finders. Distance filters are based on a simple observation: In proximity sensing, unmodeled obstacles typically produce readings that are *shorter* than the distance expected from the map. In essence, the *distance filter* selects sensor readings based on their distance relative to the distance to the closest obstacle in the map.

To be more specific, this filter removes those sensor measurements $s$ which with probability higher than $\gamma$ (this threshold is set to 0.99 in all experiments) are *shorter* than expected, and which therefore are caused by an unmodeled object (e.g. a person).

To see, let $d_1, \ldots, d_n$ be a discrete set of possible distances measured by a proximity sensor. As in Section 3.2, we denote by $P_m(d_i \mid l)$ the probability of measuring distance $d_i$ if the robot is at position $l$ and the sensor detects the closest obstacle in the map along the sensing direction. The distribution $P_m$ describes the sensor measurement *expected* from the map. As described above, this distribution is assumed to be Gaussian with mean at the distance $o_l$ to the closest obstacle along the sensing direction. The dashed line in Figure 9 represents $P_m$, for a laser range-finder and a distance $o_l$ of 230cm. We now can define the probability $P_{\text{short}}(d_i \mid l)$ that a measured distance $d_i$ is *shorter* than the expected one given the robot is at position $l$. This probability is obviously equivalent to the probability that the expected measurement $o_l$ is longer than $d_i$ given the robot is at location $l$ and thus can be computed as follows:

$$P_{\text{short}}(d_i \mid l) \quad = \quad \sum_{j>i} P_m(d_j \mid l). \tag{35}$$

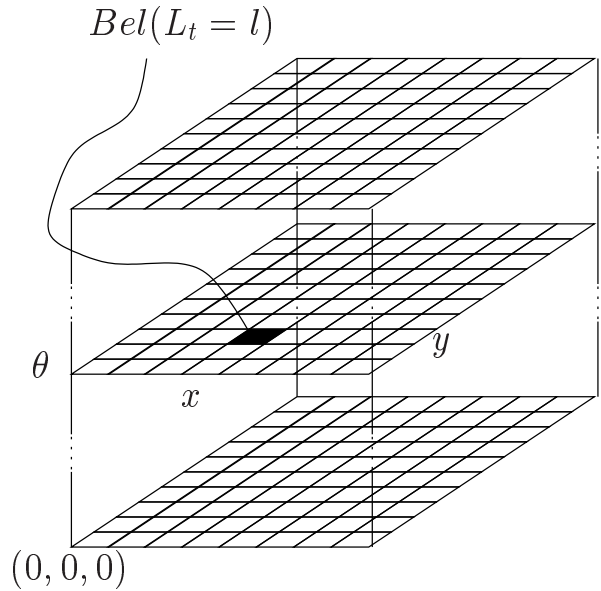

Fig. 10. Grid-based representation of the state space

In practice, however, we are interested in the probability $P_{\text{short}}(d_i)$ that $d_i$ is shorter than expected, given the complete current belief of the robot. Thus, we have to average over all possible positions of the robot:

$$P_{\text{short}}(d_i) \quad = \quad \sum_l P_{\text{short}}(d_i \mid l) Bel(L = l) \tag{36}$$

Given the distribution $P_{\text{short}}(d_i)$, we now can implement the distance filter by excluding all sensor measurements $d_i$ with $P_{\text{short}}(d_i) > \gamma$. Whereas the entropy filter filters measurements according to their effect on the belief state of the robot the distance filter selects measurements solely based on their value and regardless of their effect on the robot's certainty.

It should be noted that Fox (1998) additionally developed a *blockage* filter for proximity sensors, which is based on a probabilistic description of situations in which a sensor is blocked by an unknown obstacle. We omit this filter here since its derivation is quite complex and the resulting filter is not significantly different from the distance filter described here.

## 3.4 Grid-based Representation of the State Space

We will now return to the issue of how to represent and compute the belief distribution of the robot efficiently, describing what one might think of as the "nut and bolts" of grid-based Markov localization. Recall that to obtain accurate metric position estimates, our approach to Markov localization uses a fine-grained discretization of the state space. Here $L$ is represented by a three-dimensional, regularly spaced grid, where the spatial resolution is usually between 10cm and 40cm and the angular resolution is usually 2 or 5 degrees. Figure 10 illustrates the structure of a position probability grid. Each layer of such a grid corresponds to all possible poses of the robot with the same orientation.

While such a fine-grained approximation makes it possible to estimate the robot's position with high accuracy, an obvious disadvantage of such a fine-grained discretization lies in the huge state space which has to be maintained. For a mid-size environment of size $30 \times 30\text{m}^2$, an angular grid resolution of $2°$, and a cell size of $15 \times 15\text{cm}^2$ the state space consists of $7,200,000$ states. The basic Markov localization algorithm updates each of these states for each sensory input and each atomic movement of the robot. Current computer speed, thus, makes it impossible to update matrices of this size in real-time.

To update such state spaces efficiently, we have developed two techniques, which are described in the remainder of this section. The first method, introduced in Section 3.4.1, pre-computes the sensor model. It allows us to determine the likelihood $P(s \mid l)$ of sensor measurements by two look-up operations—instead of expensive ray tracing operations. The second optimization, described in Section 3.4.2, is a *selective update* strategy. This strategy focuses the computation, by only updating the *relevant* part of the state space. Based on these two techniques, grid-based Markov localization can be applied on-line to estimate the position of a mobile robot during its operation, using a low-cost PC.

### 3.4.1 PRE-COMPUTATION OF THE SENSOR MODEL

As described in Section 3.2, the perception model $P(s \mid l)$ for proximity sensors only depends on the distance $o_l$ to the closest obstacle in the map along the sensor beam. Based on the assumption that the map of the environment is static, our approach pre-computes and stores these distances $o_l$ for each possible robot location $l$ in the environment. Following our sensor model, we use a discretization $d_1, \ldots, d_n$ of the possible distances $o_l$. This discretization is exactly the same for the expected and the measured distances. We then store for each location $l$ only the index of the expected distance $o_l$ in a three-dimensional table. Please note that this table only needs one byte per value if 256 different values for the discretization of $o_l$ are used. The probability $P(d_i \mid o_l)$ of measuring a distance $d_i$ if the closest obstacle is at distance $o_l$ (see Figure 6) can also be pre-computed and stored in a two-dimensional lookup-table.

As a result, the probability $P(s \mid l)$ of measuring $s$ given a location $l$ can quickly be computed by two nested lookups. The first look-up retrieves the distance $o_l$ to the closest obstacle in the sensing direction given the robot is at location $l$. The second lookup is then used to get the probability $P(s \mid o_l)$. The efficient computation based on table look-ups enabled our implementation to quickly incorporate even laser-range scans that consist of up to 180 values in the overall belief state of the robot. In our experiments, the use of the look-up tables led to a speed-up-factor of 10, when compared to a computation of the distance to the closest obstacle at run-time.

### 3.4.2 SELECTIVE UPDATE

The selective update scheme is based on the observation that during global localization, the certainty of the position estimation permanently increases and the density quickly concentrates on the grid cells representing the true position of the robot. The probability of the other grid cells decreases during localization and the key idea of our optimization is to exclude unlikely cells from being updated.

For this purpose, we introduce a threshold[3] $\varepsilon$ and update only those grid cells $l$ with $Bel(L_t = l) > \varepsilon$. To allow for such a *selective* update while still maintaining a density over the *entire* state space, we approximate $P(s_t \mid l)$ for cells with $Bel(L_t = l) \le \varepsilon$ by the *a priori* probability of measuring $s_t$. This quantity, which we call $\widetilde{P}(s_t)$, is determined by averaging over all possible locations of the robot:

$$\widetilde{P}(s_t) = \sum_l P(s_t \mid l) \, P(l) \tag{37}$$

---

3. In our current implementation $\varepsilon$ is set to 1% of the *a priori* position probability.

Please note that $\widetilde{P}(s_t)$ is independent of the current belief state of the robot and can be determined beforehand. The incremental update rule for a new sensor measurement $s_t$ is changed as follows (compare Equation (9)):

$$Bel(L_t = l) \quad \longleftarrow \quad \begin{cases} \alpha_t \cdot P(s_t \mid l) \cdot Bel(L_{t-1} = l) & \text{if } Bel(L_{t-1} = l) > \varepsilon \\ \alpha_t \cdot \quad \widetilde{P}(s_t) \quad \cdot Bel(L_{t-1} = l) & \text{otherwise} \end{cases} \tag{38}$$

By multiplying $\widetilde{P}(s_t)$ into the normalization factor $\alpha_t$, we can rewrite this equation as

$$Bel(L_t = l) \quad \longleftarrow \quad \begin{cases} \tilde{\alpha}_t \cdot \frac{P(s_t \mid l)}{\widetilde{P}(s_t)} \cdot Bel(L_{t-1} = l) & \text{if } Bel(L_{t-1} = l) > \varepsilon \\ \tilde{\alpha}_t \cdot \quad\quad Bel(L_{t-1} = l) & \text{otherwise} \end{cases} \tag{39}$$

where $\tilde{\alpha}_t = \alpha_t \cdot \widetilde{P}(s_t)$.

The key advantage of the selective update scheme given in Equation (39) is that all cells with $Bel(L_{t-1} = l) \le \varepsilon$ are updated with the same value $\tilde{\alpha}_t$. In order to obtain smooth transitions between global localization and position tracking and to focus the computation on the important regions of the state space $L$, for example, in the case of ambiguities we use a partitioning of the state space. Suppose the state space $L$ is partitioned into $n$ segments or parts $\pi_1, \ldots, \pi_n$. A segment $\pi_i$ is called *active* at time $t$ if it contains locations with probability above the threshold $\varepsilon$; otherwise we call such a part *passive* because the probabilities of all cells are below the threshold. Obviously, we can keep track of the individual probabilities within a passive part $\pi_i$ by accumulating the normalization factors $\tilde{\alpha}_t$ into a value $\beta_i$. Whenever a segment $\pi_i$ becomes passive, i.e. the probabilities of all locations within $\pi_i$ no longer exceed $\varepsilon$, the normalizer $\beta_i(t)$ is initialized to 1 and subsequently updated as follows: $\beta_i(t + 1) = \tilde{\alpha}_t \cdot \beta_i(t)$. As soon as a part becomes active again, we can restore the probabilities of the individual grid cells by multiplying the probabilities of each cell with the accumulated normalizer $\beta_i(t)$. By keeping track of the robot motion since a part became passive, it suffices to incorporate the accumulated motion whenever the part becomes active again. In order to efficiently detect whether a passive part has to be activated again, we store the maximal probability $P_i^{\max}$ of all cells in the part at the time it becomes passive. Whenever $P_i^{\max} \cdot \beta_i(t)$ exceeds $\varepsilon$, the part $\pi_i$ is activated again because it contains at least one position with probability above the threshold. In our current implementation we partition the state space $L$ such that each part $\pi_i$ consists of all locations with equal orientation relative to the robot's start location.

To illustrate the effect of this selective update scheme, let us compare the update of active and passive cells on incoming sensor data. According to Equation (39), the difference lies in the ratio $P(s_t \mid l)/\widetilde{P}(s_t)$. An example of this ratio for our model of proximity sensors is depicted in Figure 11 (here, we replaced $s_t$ by a proximity measurement $d_i$). In the beginning of the localization process, all cells are active and updated according to the ratio depicted in Figure 11. The measured and expected distances for cells that do not represent the true location of the robot usually deviate significantly. Thus, the probabilities of these cells quickly fall below the threshold $\varepsilon$.

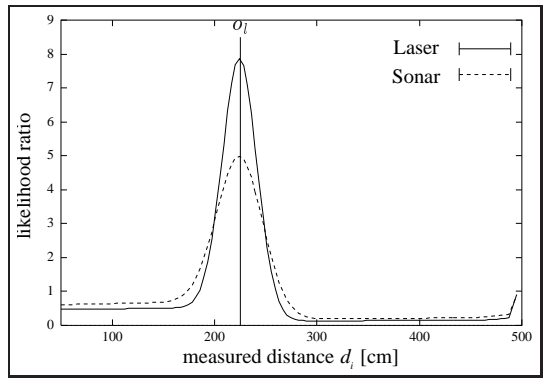

Fig. 11. Ratio $\frac{P(d_i|l)}{P(d_i)}$ for sonar and laser measurements for expected distance $o_l$ of 230cm.

Now the effect of the selective update scheme becomes obvious: Those parts of the state space that do not align well with the orientation of the environment, quickly become passive as the robot localizes itself. Consequently, only a small fraction of the state space has to be updated as soon as the robot has correctly determined its position. If, however, the position of the robot is lost, then the likelihood ratios for the distances measured at the active locations become smaller than one on average. Thus the probabilities of the active locations decrease while the normalizers $\beta_i$ of the passive parts increase until these segments are activated again. Once the true position of the robot is among the active locations, the robot is able to re-establish the correct belief.

In extensive experimental tests we did not observe evidence that the selective update scheme has a noticably negative impact on the robot's behavior. In contrast, it turned out to be highly effective, since in practice only a small fraction (generally less than 5%) of the state space has to be updated once the position of the robot has been determined correctly, and the probabilities of the active locations generally sum up to at least 0.99. Thus, the selective update scheme automatically adapts the computation time required to update the belief to the certainty of the robot. This way, our system is able to efficiently track the position of a robot once its position has been determined. Additionally, Markov localization keeps the ability to detect localization failures and to relocalize the robot. The only disadvantage lies in the fixed representation of the grid which has the undesirable effect that the memory requirement in our current implementation stays constant even if only a minor part of the state space is updated. In this context we would like to mention that recently promising techniques have been presented to overcome this disadvantage by applying alternative and dynamic representations of the state space (Burgard *et al.*, 1998b; Fox *et al.*, 1999).

## 4. Experimental Results

Our metric Markov localization technique, including both sensor filters, has been implemented and evaluated extensively in various environments. In this section we present some of the experiments carried out with the mobile robots Rhino and Minerva (see Figure 1). Rhino has a ring of 24 ultrasound sensors each with an opening angle of 15 degrees. Both, Rhino and Minerva are equipped with two laser range-finders covering a 360 degrees field of view.

The first set of experiments demonstrates the robustness of Markov localization in two real-world scenarios. In particular, it systematically evaluates the effect of the filtering techniques on the localization performance in highly dynamic environments. An additional experiment illustrates a further advantage of the filtering technique, which enables a mobile robot to reliably estimate its position even if only an outline of an office environment is given as a map.

In further experiments described in this section, we will illustrate the ability of our Markov localization technique to globally localize a mobile robot in approximate world models such as occupancy grid maps, even when using inaccurate sensors such as ultrasound sensors. Finally, we present experiments analyzing the accuracy and efficiency of grid-based Markov localization with respect to the size of the grid cells.

The experiments reported here demonstrate that Markov localization is able to globally estimate the position of a mobile robot, and to reliably keep track of it even if only an approximate model of a possibly dynamic environment is given, if the robot has a weak odometry, and if noisy sensors such as ultrasound sensors are used.

## 4.1 Long-term Experiments in Dynamic Environments

For our mobile robots Rhino and Minerva, which operated in the *Deutsches Museum Bonn* and the US-Smithsonian's *National Museum of American History*, the robustness and reliability of our Markov localization system was of utmost importance. Accurate position estimation was a crucial component, as many of the obstacles were "invisible" to the robots' sensors (such as glass cages, metal bars, staircases, and the alike). Given the estimate of the robot's position (Fox *et al.*, 1998b) integrated map information into the collision avoidance system in order to prevent the robot from colliding with obstacles that could not be detected.

Figure 12(a) shows a typical trajectory of the robot Rhino, recorded in the museum in Bonn, along with the map used for localization. The reader may notice that only the obstacles shown in black were actually used for localization; the others were either invisible or could not be detected reliably. Rhino used the entropy filter to identify sensor readings that were corrupted by the presence of people. Rhino's localization module was able to (1) globally localize the robot in the morning when the robot was switched on and (2) to reliably and accurately keep track of the robot's position. In the entire six-day deployment period, in which Rhino traveled over 18km, our approach led only to a single software-related collision, which involved an "invisible" obstacle and which was caused by a localization error that was slightly larger than a 30cm safety margin.

Figure 12(b) shows a 2km long trajectory of the robot Minerva in the National Museum of American History. Minerva used the distance filter to identify readings reflected by unmodeled objects. This filter was developed after Rhino's deployment in the museum in Bonn, based on an analysis of the localization failure reported above and in an attempt to prevent similar effects in future installations. Based on the distance filter, Minerva was able to operate reliably over a period of 13 days. During that time Minerva traveled a total of 44km with a maximum speed of 1.63m/sec.

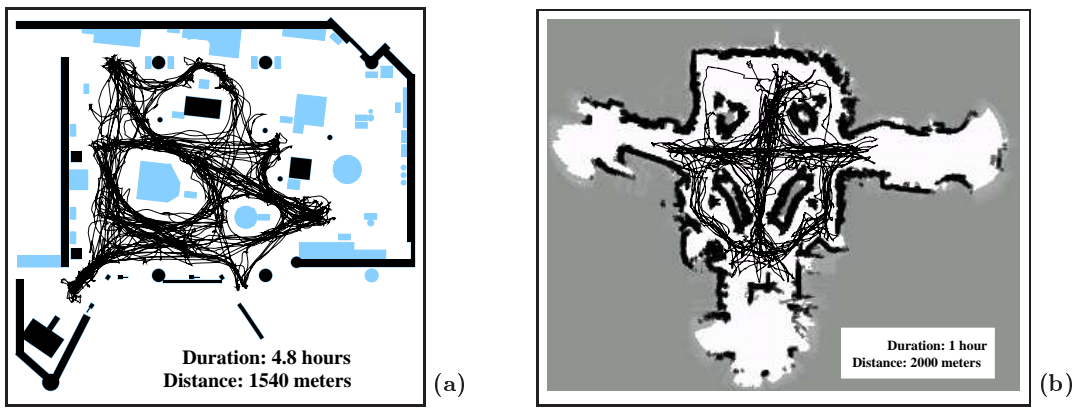

Fig. 12. Typical trajectories of (a) Rhino in the *Deutsches Museum Bonn* and (b) Minerva in the *National Museum of American History*.

Unfortunately, the evidence from the museum projects is anecdotal. Based on sensor data collected during Rhino's deployment in the museum in Bonn, we also investigated the effect of our filter techniques more systematically, and under even more extreme conditions. In particular, we were interested in the localization results a.) when the environment is densely populated (more than 50% of the sensor reading are corrupted), and b.) when the robot suffers extreme dead-reckoning errors (e.g. induced by a person carrying the robot somewhere else). Since such cases are rare, we manually inflicted such errors into the original data to analyze their effect.

### 4.1.1 DATASETS

During the experiments, we used two different datasets. These sets differ mainly in the amount of sensor noise.

a.) The first dataset was collected during 2.0 hours of robot motion, in which the robot traveled approximately 1,000 meters. This dataset was collected when the museum was closed, and the robot guided only remote Internet-visitors through the museum. The robot's top speed was 50cm/sec. Thus, this dataset was "ideal" in that the environment was only sparsely populated, and the robot moved slowly.

b.) The second dataset was recorded during a period of 4.8 hours, during which Rhino traveled approximately 1,540 meters. The path of this dataset is shown in Figure 12(a). When collecting this data, the robot operated during peak traffic hours. It was frequently faced with situations such as the one illustrated in Figure 7. The robot's top speed was 80cm/sec.

Both datasets consist of logs of odometry and laser range-finder scans, collected while the robot moved through the museum. Using the time stamps in the logs, all tests have been conducted in real-time simulation on a SUN-Ultra-Sparc 1 (177-MHz). The first dataset contained more than 32,000, and the second dataset more than 73,000 laser scans. To evaluate the different localization methods, we generated two *reference paths*, by averaging over the estimates of nine independent runs for each filter on the datasets (with small random noise added to the input data). We verified the correctness of both reference paths by visual inspection; hence, they can be taken as "ground truth."

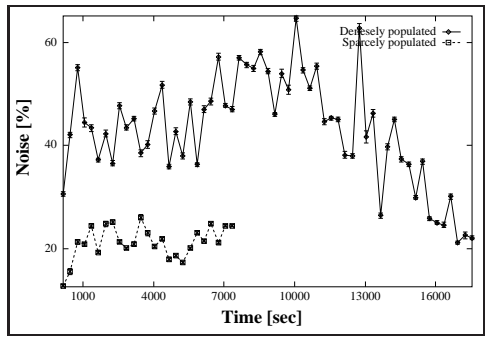

Fig. 13. Percentage of noisy sensor measurements averaged over time intervals of five minutes.

Figure 13 shows the estimated percentage of corrupted sensor readings over time for both datasets. The dashed line corresponds to the first data set, while the solid line illustrates the corruption of the second (longer) data set. In the second dataset, more than half of all measurements were corrupted for extended durations of time, as estimated by analyzing each laser reading post-facto as to whether it was significantly shorter than the distance to the next obstacle.

### 4.1.2 TRACKING THE ROBOT'S POSITION

In our first series of experiments, we were interested in comparing the ability of all three approaches—plain Markov localization without filtering, localization with the entropy filter, and localization with the distance filter—to keep track of the robot's position under normal working conditions. All three approaches tracked the robot's position in the empty museum well (first dataset), exhibiting only negligible errors in localization. The results obtained for the second, more challenging dataset, however, were quite different. In a nutshell, both filter-based approaches tracked the robot's position accurately, whereas conventional Markov localization failed frequently. Thus, had we used the latter in the museum exhibit, it would inevitably have led to a large number of collisions and other failures.

| Filter | None | | Entropy | | Distance | |
|--------|------|------|---------|------|----------|------|
| failures$_I$ [%] | 1.6 | ± 0.4 | 0.9 | ± 0.4 | 0.0 | ± 0.0 |
| failures$_{II}$ [%] | 26.8 | ± 2.4 | 1.1 | ± 0.3 | 1.2 | ± 0.7 |

Table 2: Ability to track the robot's position.

Table 2 summarizes the results obtained for the different approaches in this tracking experiment. The first row of Table 2 provides the percentage of failures for the different filters on the first dataset (error values represent 95% confidence intervals). Position estimates were considered a "failure" if the estimated location of the robot deviated from the reference path by more than 45cm for at least 20 seconds. The percentage is measured in time during which the position was lost, relative to the total time of the dataset.

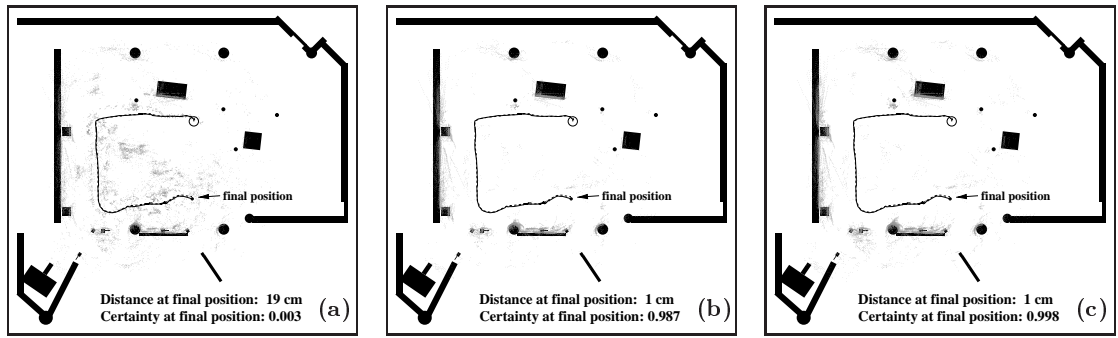

Fig. 14. Estimated and real paths of the robot along with endpoints of incorporated sensor measurements using (a) no filter, (b) entropy filter, and (c) distance filter.

As can be seen here, all three approaches work well, and the distance filter provides the best performance. The second row provides the failures on the second dataset. While plain Markov localization failed in 26.8% of the overall time, both filter techniques show almost equal results with a failure of less than 2%. Thus, the two filter techniques are robust in highly dynamic environments, plain Markov localization is prone to fail.

To shed light onto the question as to why Markov localization performs so poorly when compared to the filter algorithms, we analyzed the sensor readings that each method used during the localization task. Figure 14 shows, for a a small fraction of the data, the measurements incorporated into the robot's belief by the three different approaches. Shown there are the end points of the sensor measurements used for localization *relative to the positions on the reference path*. Obviously, both filter approaches manage to focus their attention on the "correct" sensor measurements, whereas plain Markov localization incorporates massive amounts of corrupted (misleading) measurements. As also illustrated by Figure 14, both filter-based approaches produce more accurate results with a higher certainty in the correct position.

### 4.1.3 RECOVERY FROM EXTREME LOCALIZATION FAILURES

We conjecture that a key advantage of the original Markov localization technique lies in its ability to *recover* from extreme localization failures. Re-localization after a failure is often more difficult than global localization from scratch, since the robot starts with a belief that is centered at a completely wrong position. Since the filtering techniques use the current belief to select the readings that are incorporated, it is not clear that they still maintain the ability to recover from global localization failures.

To analyze the behavior of the filters under such extreme conditions, we carried out a series of experiments during which we manually introduced such failures into the data to test the robustness of these methods in the extreme. More specifically, we "tele-ported" the robot at random points in time to other locations. Technically, this was done by changing the robot's orientation by 180±90 degree and shifting it by 0±100cm, without letting the robot know. These perturbations were introduced randomly, with a probability of 0.005 per meter of robot motion. Obviously, such incidents make the robot lose track of its position. Each method was tested on 20 differently corrupted versions of both datasets. This resulted in a total of more than 50 position failures in each dataset. For each of these failures we measured the time until the methods re-localized the robot correctly. Re-Localization was assumed to have succeeded if the distance between the estimated position and the reference path was smaller than 45cm for more than 10 seconds.

| Filter | None | | Entropy | | Distance | |
|---|---|---|---|---|---|---|
| Dataset I | | | | | | |
| $\overline{t_{\text{rec}}}$ [sec] | 237 | ± 27 | 1779 | ± 548 | 188 | ± 30 |
| failures [%] | 10.2 | ± 1.8 | 45.6 | ± 7.1 | 6.8 | ± 1.6 |
| Dataset II | | | | | | |
| $\overline{t_{\text{rec}}}$ [sec] | 269 | ± 60 | 1310 | ± 904 | 235 | ± 46 |
| failures [%] | 39.5 | ± 5.1 | 72.8 | ± 7.3 | 7.8 | ± 1.9 |

Table 3: Summary of recovery experiments.

Table 3 provides re-localization results for the various methods, based on the two different datasets. Here $\overline{t_{\text{rec}}}$ represents the average time in seconds needed to recover from a localization error. The results are remarkably different from the results obtained under normal operational conditions. Both conventional Markov localization and the technique using distance filters are relatively efficient in recovering from extreme positioning errors in the first dataset, whereas the entropy filter-based approach is an order of magnitude less efficient (see first row in Table 3). The unsatisfactory performance of the entropy filter in this experiment is due to the fact that it disregards all sensor measurements that do not confirm the belief of the robot. While this procedure is reasonable when the belief is correct, it prevents the robot from detecting localization failures. The percentage of time when the position of the robot was lost in the entire run is given in the second row of the table. Please note that this percentage includes both, failures due to manually introduced perturbations and tracking failures. Again, the distance filter is slightly better than the approach without filter, while the entropy filter performs poorly. The average times $\overline{t_{\text{rec}}}$ to recover from failures on the second dataset are similar to those in the first dataset. The bottom row in Table 3 provides the percentage of failures for this more difficult dataset. Here the distance filter-based approach performs significantly better than both other approaches, since it is able to quickly recover from localization failures and to reliably track the robot's position.

The results illustrate that despite the fact that sensor readings are processed selectively, the distance filter-based technique recovers as efficiently from extreme localization errors as the conventional Markov approach.

## 4.2 Localization in Incomplete Maps

A further advantage of the filtering techniques is that Markov localization does not require a detailed map of the environment. Instead, it suffices to provide only an outline which merely includes the aspects of the world which are static. Figure 15(a) shows a ground plan of our department building, which contains only the walls of the university building. The complete map, including all movable objects such as tables and chairs, is shown in Figure 19. The two Figures 15(b) and 15(c) illustrate how the distance filter typically behaves when tracking the robot's position in such a sparse map of the environment. Filtered readings are shown in grey, and the incorporated sensor readings are shown in black. Obviously, the filter focuses on the known aspects of the map and ignores all objects (such as desks, chairs, doors and tables) which are not contained in the outline. Fox (1998) describes more systematic experiments supporting our belief that Markov localization in combination with the distance filter is able to accurately localize mobile robots even when relying only on an outline of the environment.

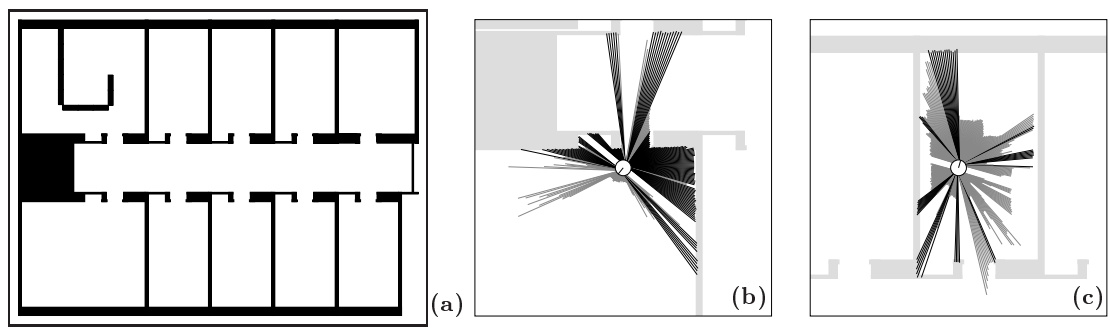

Fig. 15. (a) Outline of the office environment and (b,c) examples of filtered (grey) and incorporated (black) sensor readings using the distance filter.

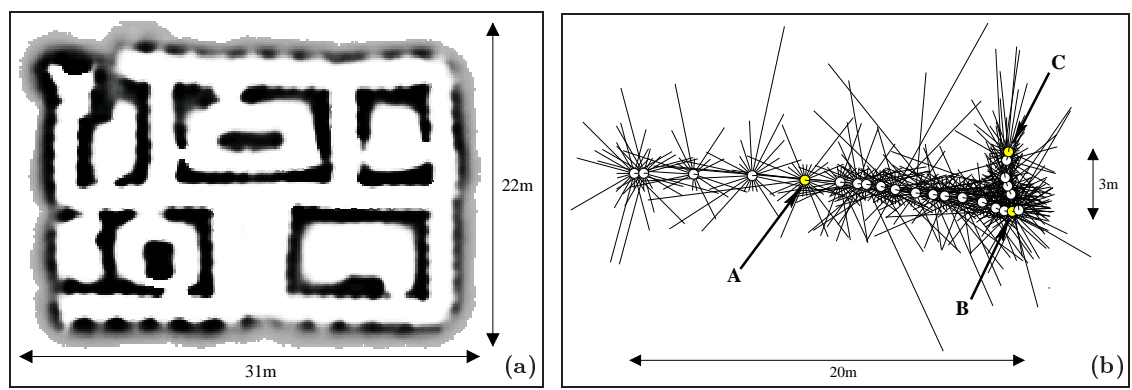

Fig. 16. (a) Occupancy grid map of the 1994 AAAI mobile robot competition arena. (b) Trajectory of the robot and ultrasound measurements used to globally localize the robot in this map.

## 4.3 Localization in Occupancy Grid Maps Using Sonar

The next experiment described here is carried out based on data collected with the mobile robot Rhino during the 1994 AAAI mobile robot competition (Simmons, 1995). Figure 16(a) shows an occupancy grid map (Moravec & Elfes, 1985; Moravec, 1988) of the environment, constructed with the techniques described in (Thrun *et al.*, 1998a; Thrun, 1998b). The size of the map is $31 \times 22m^2$, and the grid resolution is 15cm.

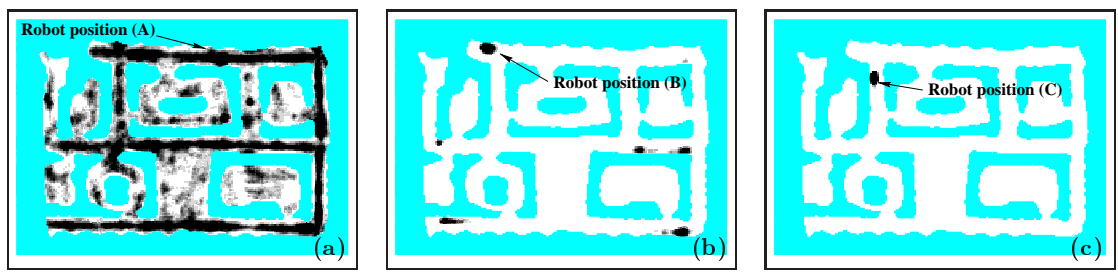

Fig. 17. Density plots after incorporating 5, 18, and 24 sonar scans (darker positions are more likely).

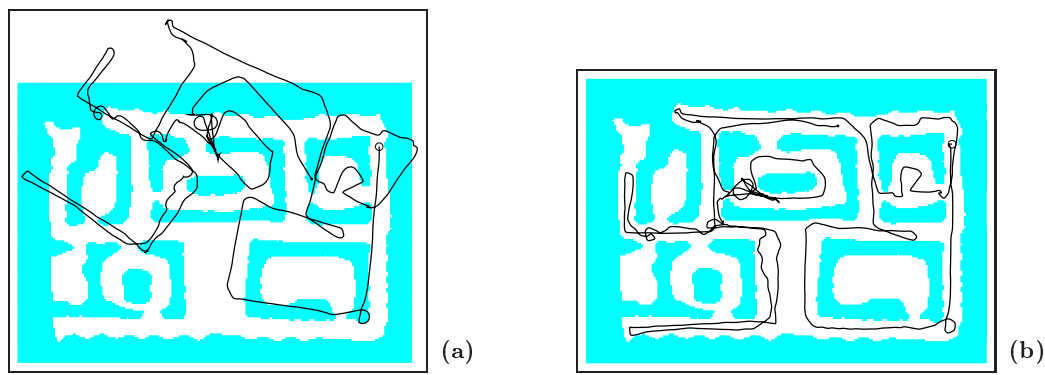

Fig. 18. Odometry information and corrected path of the robot.

Figure 16(b) shows a trajectory of the robot along with measurements of the 24 ultrasound sensors obtained as the robot moved through the competition arena. Here we use this sensor information to globally localize the robot from scratch. The time required to process this data on a 400MHz Pentium II is 80 seconds, using a position probability grid with an angular resolution of 3 degrees. Please note that this is exactly the time needed by the robot to traverse this trajectory; thus, our approach works in real-time. Figure 16(b) also marks positions of the robot after perceiving 5 (A), 18 (B), and 24 (C) sensor sweeps. The belief states during global localization at these three points in time are illustrated in Figure 17.

The figures show the belief of the robot projected onto the $\langle x, y \rangle$-plane by plotting for each $\langle x, y \rangle$-position the maximum probability over all possible orientations. More likely positions are darker and for illustration purposes, Figures 17(a) and 17(b) use a logarithmic scale in intensity. Figure 17(a) shows the belief state after integrating 5 sensor sweeps (see also position A in Figure 16(b)). At this point in time, all the robot knows is that it is in one of the corridors of the environment. After integrating 18 sweeps of the ultrasound sensors, the robot is almost certain that it is at the end of a corridor (compare position B in Figures 16(b) and 17(b)). A short time later, after turning left and integrating six more sweeps of the ultrasound ring, the robot has determined its position uniquely. This is represented by the unique peak containing 99% of the whole probability mass in Figure 17(c).

Figure 18 illustrates the ability of Markov localization to correct accumulated dead-reckoning errors by matching ultrasound data with occupancy grid maps. Figure 18(a) shows a typical 240m long trajectory, measured by Rhino's wheel-encoders in the 1994

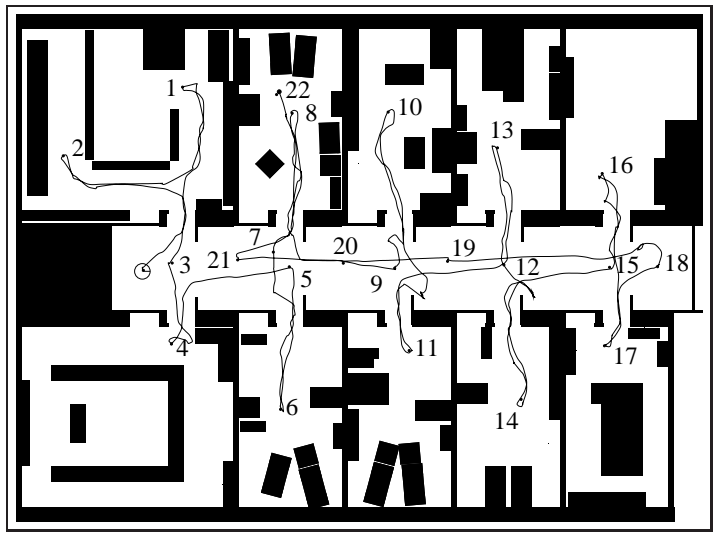

Fig. 19. Path of the robot and reference positions

AAAI mobile robot competition arena. Obviously, the rotational error of the odometry quickly increases. Already after traveling 40m, the accumulated error in the orientation (raw odometry) is about 50 degrees. Figure 18(b) shows the path of the robot estimated by Markov localization, which is significantly more correct.

## 4.4 Precision and Performance

We will now describe experiments aimed at characterizing the precision of position estimates. Our experiments also characterize the time needed for global localization in relation to the size of the grid cells. Figure 19 shows a path of the robot Rhino in the Computer Science Department's building at the University of Bonn. This path includes 22 reference positions, where the true position of the robot was determined using the scan matching technique presented in (Gutmann & Schlegel, 1996; Lu & Milios, 1994). All data recorded during this run were split into four disjoint traces of the sensor data. Each of these different traces contained the full length of the path, but only every fourth sensor reading which was sufficient to test the localization performance.

Figure 20(a) shows the localization error averaged over the four runs and all reference positions. The error was determined for different sizes of grid cells, using a laser range-finder or ultrasound sensors. These results demonstrate (1) that the average localization error for both sensors is generally below the cell size and (2) that laser range-finders provide a significantly higher accuracy than ultrasound sensors. When using the laser range-finder at a spatial resolution of 4cm, the average positioning error can even be reduced to 3.5cm.

Figure 20(b) shows the average CPU-time needed to globally localize the robot as a function of the size of the grid cells. The values represent the computation time needed on a 266MHz Pentium II for global localization on the path between the starting point and position 1. In this experiment, we used a fixed angular resolution of four degrees. In the case of 64cm cell size, the average localization time is approximately 2.2 seconds.

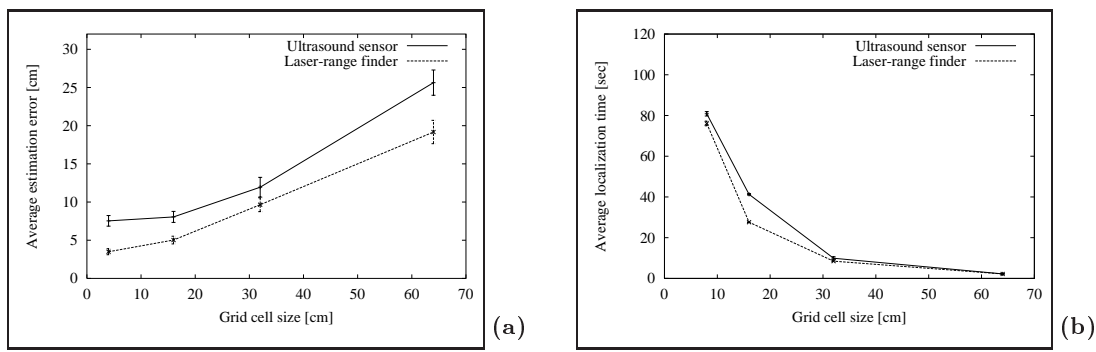

Fig. 20. (a) Average localization error and (b) average CPU-time needed for global localization time both for ultrasound sensors and laser range-finder depending on the grid resolution.

Of course, the effective time needed for global localization in practice highly depends on the structure of the environment and the amount of information gathered on the path of the robot. For example, due to the symmetry of the corridor of this office environment, the robot is not able to localize itself unless it enters a room. The reader may notice that recently, we developed a decision-theoretic method for actively guiding the robot to places which allow it to resolve ambiguities during global localization (Fox *et al.*, 1998a; Fox, 1998). Based on this method, the localization process becomes more efficient, especially in office environments with a lot of indistinguishable places as, for example, long corridors.

The experiments described above demonstrate that our metric variant of Markov localization is able to efficiently estimate the position of a mobile robot in dynamic environments. It furthermore can deal with approximate models of the environment such as occupancy grid maps or rough outline maps. Finally, it is able to efficiently and accurately estimate the position of a mobile robot even if ultrasound sensors are used.

## 5. Related Work

Most of the techniques for mobile robot localization in the literature belong to the class of local approaches or tracking techniques, which are designed to compensate odometric error occurring during navigation. They assume that the initial position of the robot is known (see Borenstein *et al.* 1996 for a comprehensive overview). For example, Weiß *et al.* (1994) store angle histograms constructed out of laser range-finder scans taken at different locations in the environment. The position and orientation of the robot are calculated by maximizing the correlation between the stored histograms and laser range-scans obtained while the robot moves through the environment. The estimated position, together with the odometry information, is then used to predict the position of the robot and to select the histogram used for the next match. Yamauchi (1996) and Schulz *et al.* (1999) apply a similar technique, but they use hill-climbing to match local maps built from ultrasound sensors into a global occupancy grid map. As in the approach by Weiß *et al.* (1994), the location of the robot is represented by the position yielding the best match. These techniques, in contrast to Markov localization, do not represent the uncertainty of the robot in its current belief and therefore cannot deal appropriately with globally ambiguous situations.

A popular probabilistic framework for position tracking are *Kalman filters* (Maybeck, 1990; Smith *et al.*, 1990), a signal processing technique introduced by Kalman (1960). As mentioned in Section 2.4, Kalman filter-based methods represent their belief of the robot's position by a unimodal Gaussian distribution over the three-dimensional state-space of the robot. The mode of this distribution yields the current position of the robot, and the variance represents the robot's uncertainty. Whenever the robot moves, the Gaussian is shifted according to the distance measured by the robot's odometry. Simultaneously, the variance of the Gaussian is increased according to the model of the robot's odometry. New sensory input is incorporated into the position estimation by matching the percepts with the world model.

Existing applications of Kalman filtering to position estimation for mobile robots are similar in how they model the motion of the robot. They differ mostly in how they update the Gaussian according to new sensory input. Leonard and Durrant-Whyte (1991) match beacons extracted from sonar scans with beacons predicted from a geometric map of the environment. These beacons consist of planes, cylinders, and corners. To update the current estimate of the robot's position, Cox (1991) matches distances measured by infrared sensors with a line segment description of the environment. Schiele and Crowley (1994) compare different strategies to track the robot's position based on occupancy grid maps and ultrasonic sensors. They show that matching local occupancy grid maps with a global grid map results in a similar localization performance as if the matching is based on features that are extracted from both maps. Shaffer *et al.* (1992) compare the robustness of two different matching techniques with different sources of noise. They suggest a combination of map-matching and feature-based techniques in order to inherit the benefits of both. Lu and Milios (1994,1997b) and Gutmann and Schlegel (1996) use a scan-matching technique to precisely estimate the position of the robot based on laser range-finder scans and learned models of the environment. Arras and Vestli (1998) use a similar technique to compute the position of the robot with a very high accuracy. All these variants, however, rest on the assumption that the position of the robot can be represented by a single Gaussian distribution. The advantage of Kalman filter-based techniques lies in their efficiency and in the high accuracy that can be obtained. The restriction to a unimodal Gaussian distribution, however, is prone to fail if the position of a robot has to be estimated from scratch, i.e. without knowledge about the starting position of the robot. Furthermore, these technique are typically unable to recover from localization failures. Recently, Jensfelt and Kristensen (1999) introduced an approach based on multiple hypothesis tracking, which allows to model multi-modal probability distributions as they occur during global localization.

Markov localization, which has been employed successfully in several variants (Nourbakhsh *et al.*, 1995; Simmons & Koenig, 1995; Kaelbling *et al.*, 1996; Burgard *et al.*, 1996; Hertzberg & Kirchner, 1996; Koenig & Simmons, 1998; Oore *et al.*, 1997; Thrun, 1998a), overcomes the disadvantage of Kalman filter based techniques. The different variants of this technique can be roughly distinguished by the type of discretization used for the representation of the state space. Nourbakhsh *et al.* (1995), Simmons and Koenig (1995), and Kaelbling *et al.* (1996) use Markov localization for landmark-based navigation, and the state space is organized according to the topological structure of the environment. Here nodes of the topological graph correspond to distinctive places in hallways such as openings or junctions and the connections between these places. Possible observations of the robot are, for example, hallway intersections. The advantage of these approaches is that they can represent ambiguous situations and thus are in principle able to globally localize a robot. Furthermore, the coarse discretization of the environment results in relatively small state spaces that can be maintained efficiently. The topological representations have the disadvantage that they provide only coarse information about the robot's position and that they rely on the definition of abstract features that can be extracted from the sensor information. The approaches typically make strong assumptions about the nature of the environments. Nourbakhsh *et al.* (1995), Simmons and Koenig (1995), and Kaelbling *et al.* (1996), for example, only consider four possible headings for the robot position assuming that the corridors in the environment are orthogonal to each other.

Our method uses instead a fine-grained, grid-based discretization of the state space. The advantage of this approach compared to the Kalman filter based techniques comes from the ability to represent more complex probability distributions. In a recent experimental comparison to the technique introduced by Lu and Milios (1994) and Gutmann and Schlegel (1996), we found that Kalman filter based tracking techniques provide highly accurate position estimates but are less robust, since they lack the ability to globally localize the robot and to recover from localization errors (Gutmann *et al.*, 1998). In contrast to the topological implementations of Markov localization, our approach provides accurate position estimates and can be applied even in highly unstructured environments (Burgard *et al.*, 1998a; Thrun *et al.*, 1999). Using the selective update scheme, our technique is able to efficiently keep track of the robot's position once it has been determined. It also allows the robot to recover from localization failures.

Finally, the vast majority of existing approaches to localization differ from ours in that they address localization in static environments. Therefore, these methods are prone to fail in highly dynamic environments in which, for example, large crowds of people surround the robot (Fox *et al.*, 1998c). However, dynamic approaches have great practical importance, and many envisioned application domains of service robots involve people and populated environments.

## 6. Discussion

In this paper we presented a metric variant of Markov localization, as a robust technique for estimating the position of a mobile robot in dynamic environments. The key idea of Markov localization is to maintain a probability density over the whole state space of the robot relative to its environment. This density is updated whenever new sensory input is received and whenever the robot moves. Metric Markov localization represents the state space using fine-grained, metric grids. Our approach employs efficient, selective update algorithms to update the robot's belief in real-time. It uses filtering to cope with dynamic environments, making our approach applicable to a wide range of target applications.

In contrast to previous approaches to Markov localization, our method uses a fine-grained discretization of the state space. This allows us to compute accurate position estimates and to incorporate raw sensory input into the belief. As a result, our system can exploit arbitrary features of the environment. Additionally, our approach can be applied in arbitrary unstructured environments and does not rely on an orthogonality assumption or similar assumptions of the existence of certain landmarks, as most other approaches to Markov localization do.

The majority of the localization approaches developed so far assume that the world is static and that the state of the robot is the only changing aspect of the world. To be able to localize a mobile robot even in dynamic and densely populated environments, we developed a technique for filtering sensor measurements which are corrupted due to the presence of people or other objects not contained in the robot's model of the environment.

To efficiently update the huge state spaces resulting from the grid-based discretization, we developed two different techniques. First, we use look-up operations to efficiently compute the quantities necessary to update the belief of the robot given new sensory input. Second, we apply the selective update scheme which focuses the computation on the relevant parts of the state space. As a result, even large belief states can be updated in real-time.

Our technique has been implemented and evaluated in several real-world experiments at different sites. Recently we deployed the mobile robots Rhino in the *Deutsches Museum Bonn*, Germany, and Minerva in the Smithsonian's *National Museum of American History*, Washington, DC, as interactive museum tour-guides. During these deployments, our Markov localization technique reliably estimated the position of the robots over long periods of time, despite the fact that both robots were permanently surrounded by visitors which produced large amounts of false readings for the proximity sensors of the robots. The accuracy of grid-based Markov localization turned out to be crucial to avoid even such obstacles that could not be sensed by the robot's sensors. This has been accomplished by integrating map information into the collision avoidance system (Fox *et al.*, 1998b).

Despite these encouraging results, several aspects warrant future research. A key disadvantage of our current implementation of Markov localization lies in the fixed discretization of the state space, which is always kept in main memory. To scale up to truly large environments, it seems inevitable that one needs variable-resolution representations of the state space, such as as the one suggested in (Burgard *et al.*, 1997; 1998b; Gutmann *et al.*, 1998). Alternatively, one could use Monte-Carlo based representations of the state space as described in (Fox *et al.*, 1999). Here, the robot's belief is represented by samples that concentrate on the most likely parts of the state space.

## Acknowledgment

The authors would like to thank the research group for autonomous intelligent systems at the University of Bonn for fruitful discussions, useful suggestions and comments, especially Daniel Hennig and Andreas Derr. We would also like to thank the members of CMU's Robot Learning Lab for many inspiring discussions. Finally, we would like to thank the staff of the Deutsches Museum Bonn and the National Museum of American History for their enthusiasm and their willingness to expose their visitors to one of our mobile robots.

This research is sponsored in part by NSF (CAREER Award IIS-9876136) and DARPA via TACOM (contract number DAAE07-98-C-L032), and Rome Labs (contract number F30602-98-2-0137), which is gratefully acknowledged. The views and conclusions contained in this document are those of the authors and should not be interpreted as necessarily representing official policies or endorsements, either expressed or implied, of NSF, DARPA, TACOM, Rome Labs, or the United States Government.